\documentclass[pmlr]{jmlr}

\makeatletter
\def\set@curr@file#1{\def\@curr@file{#1}} 
\makeatother

\usepackage{longtable}

\usepackage{booktabs}
\usepackage[load-configurations=version-1]{siunitx} 

\theorembodyfont{\upshape}
\theoremheaderfont{\scshape}
\theorempostheader{:}
\theoremsep{\newline}

\jmlrvolume{}
\jmlryear{}
\jmlrworkshop{}

\def\longtitle{Interpretable Factorization for Neural Network ECG Models}
\def\shorttitle{Neural Component Analysis}
\title[\shorttitle{}]{\longtitle{}}

\author{\Name{Christopher Snyder}
       \Email{22csynder@gmail.com}\\ 
       \addr Department of Biomedical Engineering\\The University of Texas at Austin\\
       Austin, Texas USA 
       \AND
       \Name{Sriram Vishwanath}
       \Email{sriram@austin.utexas.edu}\\ 
       \addr Department of Electrical and Computer Engineering\\The University of Texas at Austin\\
       Austin, Texas USA}

\usepackage{verbatim}
\usepackage[x11names]{xcolor}

\def\x{x}

\def\Wmat{W}
\newcommand{\WeightMatrix}[1]{\Wmat^{(#1)}}
\newcommand{\W}[1]{\WeightMatrix{#1}}
\newcommand{\bs}{b}

\newcommand{\B}[1]{\bs^{(#1)}}

\def\relu{R}

\def\net{\mathcal{N}}

\usepackage{soul,xcolor}
\setstcolor{red}

\def\bbR{\mathbb{R}}

\usepackage{bbm}

\def\Pi{P^{I}}

\def\Mm{MinMax}
\def\MmRep{\Mm{}-Representation} 

\def\MmChar{Character}
\def\MmFactor{\MmChar{} Function}    
\def\MmSpace{Attribute Space}
\def\MmConcept{model concept}
\def\MmFun{\Psi}
\def\MmPart{\phi}

\def\posthoc{post hoc}
\def\blackbox{black box}
\begin{document}

\maketitle

\begin{abstract}
The ability of deep learning (DL) to improve the practice of medicine and its clinical outcomes faces a looming obstacle: model interpretation. Without description of how outputs are generated, a collaborating physician can neither resolve when the model's conclusions are in conflict with his or her own, nor learn to anticipate model behavior. Current research aims to interpret networks that diagnose ECG recordings, which has great potential impact as recordings become more personalized and widely deployed. A generalizable impact beyond ECGs lies in the ability to provide a rich test-bed for the development of interpretive techniques in medicine. Interpretive techniques for Deep Neural Networks (DNNs), however, tend to be heuristic and observational in nature, lacking the mathematical rigor one might expect in the analysis of math equations. The motivation of this paper is to offer a third option, a scientific approach. 
We treat the model output itself as a phenomenon to be explained through component parts and equations governing their behavior. We argue that these component parts should \emph{also} be ``black boxes"
--additional targets to interpret heuristically with clear functional connection to the original. 
We show how to rigorously factor a DNN into a hierarchical equation consisting of black box variables. 
This is not a subdivision into physical parts, like an organism into its cells; it is but one choice of an equation into a collection of abstract functions. Yet, for DNNs trained to identify normal ECG waveforms on PhysioNet 2017 Challenge data, we demonstrate this choice yields interpretable component models identified with visual composite sketches of ECG samples in corresponding input regions. Moreover, the recursion distills this interpretation: additional factorization of component black boxes corresponds to ECG partitions that are more morphologically pure. 
\end{abstract}

\section{Introduction}
Deep Neural Networks (DNNs) are a class of general purpose, or \blackbox{}, models that have immense promise for revolutionizing clinical care \citep{Porumb2020, Minchole2019}. Yet, widespread adoption of these high performance \blackbox{} models has been impeded by decreased understanding of patient level outputs. 
Although interpretability of these models is a burgeoning area of study, the existing methods of interpreting DNNs for medical predictions still show room for improvement \cite{Sethi2020}. 
For DNNs, these methods are generally applied to trained models in a \posthoc{}, unprincipled manner. The lack of rigor makes it difficult to predict when they can be relied upon for clinical diagnosis.
With this work, we extend DL in the healthcare space by applying our \posthoc{} interpretability method to ECG classification. Numerous studies have shown incremental improvement on the performance of automated DNN ECG classification, so our main focus is to improve the interpretation of ECG classification outputs. By breaking down a trained neural network into simplified component parts, a causal understanding of why the network predicts a certain outcome can be formed. The ability to quickly interpret why the network outputs its prediction will improve the diagnosis and enhance clinical understanding of the problem.

While early machine learning methods sought to encode logic about the world by hand, it was discovered that many relationships, even ones that humans found trivial, were difficult to interpret and translate explicitly into code. The issue at hand is that we find these \blackbox{} methods, initially designed to learn formulae too difficult to codify directly, now too complicated to interpret directly. In this case, model explanation becomes quite literally a phenomenological study--one that seeks descriptive generalizations of DNN behavior from (\posthoc{}) experiment and observation. We are simply pointing out this is the scientific process, adapted to explaining phenomena of math instead of nature. Hence, this strange new challenge in data science of providing high level explanations for models we can \textit{define} but struggle to \textit{describe} may be a situation with which clinicians are more familiar. In fact, we can motivate our approach to model interpretation through medical analogy, as indicated in the following section. 

\paragraph{Generalizable Insights about Machine Learning in the Context of Healthcare} 

\begin{itemize}
    \item A counter-intuitive but useful first step to \blackbox{} model interpretation is increasing the number of \blackbox{} models requiring interpretation. In medicine, this process is familiar. All of the properties we care about, like the output of neural networks, are emergent features arising from repeated composition of very simple rules. Somehow, a very simple differential equation is sufficient to predict the emergence of lymphoma from DNA sequences using physical laws alone. The challenge of interpreting neural networks is like interpreting this functional relationship without knowing in advance about all the structure in between. Without knowledge of ``cells", ``lymph nodes", even certain ``viruses", we would simply lack the vocabulary to provide useful interpretation. This is what is currently being attempted. We must instead try to discover this structure, building the interpretation of the whole model on our best understanding of its parts. We propose one such method for for neural networks classifying ECG waveforms.
    \item When we apply this method experimentally, there are two observations of fundamental interest: 
    \begin{enumerate}
        \item Factorization \textit{as functions} of interpretable DNN models results in component functions that are also interpretable, mapping to abstractions that are components of an explanation. In principle, they could be any strange functions satisfying the same equations. 
        \item Repeated factorization produces \textit{cleaner} interpretations: Not only do they remain interpretable, they become easier to interpret. 
        Surely, none of this is guaranteed in the general case, making it important to study which clinical settings qualify.
    \end{enumerate}
    
\end{itemize}

\section{Related Work}

Extensive research has demonstrated the practicality of ECG analysis for various use cases in machine learning (ML) for healthcare. DL has been shown to outperform existing risk metrics for cardiovascular death, as demonstrated by the analysis of long term patient ECGs with a DNN \citep{Shanmugam2018}.
\citet{Gupta2019} 
finds the most expressive combination of ECG leads, testing combinations from $15$ leads and training a convolutional neural network (CNN) for state of the art performance in myocardial infarction detection. 
Accurate performance has also been achieved for single-lead ECG data; 
\citet{Yildirim2018} 
use CNNs on 10 second ECG fragments for the classification of seventeen types of cardiac arrhythmia. 
Similar DNNs have also been shown to outperform board-certified cardiologists in its sensitivity when classifying single-lead ECGs into 12 rhythm classes. 
\citep{Hannun2019}. 







Atrial fibrillation classification in the PhysioNet 2017 Challenge closely resembles our focus for research with ECG signals. Our work extends the ideas present in 
\cite{Goodfellow2018}, 
who created a high performance model and interpreted its behavior with class activation maps (CAMs). The CAMs visualize typical behavior for the three target labels of an ECG signal: normal rhythm, atrial fibrillation, or other. In order to use CAMs, they first modify a top performing model developed for the original challenge. By removing many of the original max pooling layers, their newer model contains a higher temporal resolution at the layer from which they extract the CAMs. Without this architecture-specific change, the output of the mapping would not be very informative. For DNNs, most of the \posthoc{} methods still require extensive tuning to develop a reasonable understanding of their decision-making
\citep{Sethi2020}. 
With visual data, these methods provide quick assessment of high-dimensional data but they often highlight fuzzy areas of the input with little pathological importance. 



Outside of healthcare, similar visualizations are being used to characterize large networks with intuitive interfaces 
\citep{Hohman2020a}. 
We aim to further contribute to interpretable visualizations by applying our method to ECG data. By visualizing the component parts of a classification DNN, we aim to find structure in its intermediate decisions that align with our current diagnostic procedure for ECG signals. The ability to derive phenotypes from machine learning algorithms is unexplored in the clinical landscape, though the importance of explainability and interpretability  are  becoming  crucial  for  machine  learning  to  be  used  in  the clinical setting 
\citep{Tonekaboni2019}. 
Instead of applying an algorithm to each input, we break down the model into component features that explain the output for clustered input types. For ECG signals, these clusters are directly inspectable and offer insight into possible phenotypes the model deduces, which further contribute to the clinical understanding of the problem. 




\section{\MmRep{} as a Tool for Interpretation}
When we train a DNN model to fit a data pattern, what related high-level concepts does the model learn in the process? Of course, this question makes no sense as stated. The model is just a sequence of math symbols with rules for combination. It does not ``know" about abstraction. Yet, in this section we will motivate and propose a mathematically rigorous theory that makes sense of the initial question. We develop formulae relating model outputs to ``model concepts". 

\subsection{Theory: Motivation, Definitions}\label{sec:motivation}
For exposition and experimentation, we will use binary ECG classification using a DNN model as a running example. We will use $x$ to denote the input, which is a numeric representation of the ECG signal, and $\net$ to be a trained DNN model with scalar output $\net(x)$. In this context, ``trained" means that on some example inputs, the set of positive predictions where $\net(x)>0$ \textit{more or less} coincides with the cases where ``$x$ is a normal ECG". Keeping with the set notation, understanding how our model will perform in the ``real-world" is equivalent to understanding the domain of the same set of positive predictions extended now over \textit{all} possible inputs, $\{\text{all ECGs }x \text{ such that }  \net(x)>0\}$. In this notation, a valid ``model explanation" is simply a concise description of this set for humans. 

One possible example model explanation might be that $\net(x)>0$ (returns normal) if ``No ST elevation" ``AND" ``QT elongation". Here, we would consider "No ST elevation" and ``no QT elongation" to both be concepts interpretable to humans since cardiologists can readily evaluate which if either apply to a particular ECG. We also see each concept has a corresponding input set, e.g., $\{x | x\text{ has ST elevation}\}$, 
and that our abstract interpretation is really saying mathematically that $\{x | \net>0\}$ is an intersection of two sets corresponding to the familiar concepts ``ST elevation". In fact all of the ways we combine concepts (AND, OR, NOT, etc.) all have corresponding set  operations ($\cap$, $\cup$, complement, etc.). 

Therefore, we consider the task of classification model interpretation to be equivalent to finding a combination of \textit{interpretable sets} using set operations that approximates sufficiently $\{x|\net(x)>0\}$. Here, a concept is just a subset of inputs defined by a property, and that concept is interpretable if a human can reasonably decide whether that property applies. Now, \textit{finding} such a vocabulary of concepts and set operations starting from a given model is in fact \textit{fitting a second model} this time over concept combinations, with under-fitting and over-fitting failure modes. This is a difficult problem under intense study. 

Instead of tackling this problem directly, what we propose instead is a method for generating intermediate targets for interpretation, 
$\{x|\MmPart{}(x)_1>0\}$,
$\{x|\MmPart{}(x)_2>0\}$, whose intermediate interpretation is related to $\{x | \net>0\}$ through a closed form, interpretable equation. To discuss this, we need to introduce a definition.

\begin{definition}\label{def:MmRep}\MmRep{}: \\
Let integer $k>0$ be arbitrary and $\net, \MmPart{}_1,\ldots,\MmPart{}_k$ be a real valued functions of input $x$. We call $\MmFun{}:\bbR^k\mapsto\bbR$ a \MmRep{} if through composition it is generated by a (finite) number of compositions of Max and Min functions applied to subsets of the $k$ scalar inputs. If also $\MmFun{}(\MmPart{}_1(x),\ldots,\MmPart{}_k(x))=\net(x)$, then we call $\MmFun{}$ a \MmRep{} of $\net$ with \MmFactor{}s  $\MmPart{}_1,\ldots,\MmPart{}_k$
\end{definition}

The benefit of interpreting \MmRep{}s of \MmFactor{}s is that they map directly OR/AND combinations of interpretations of \MmFactor{}s. Firstly, this avoids introducing approximation or heuristic in this step. The nature of understanding DNNs probably unavoidably involves both subjectivity \textit{and} approximation at some stage, but it's helpful to know that this step can be relied upon when analyzing how things go wrong. Secondly, and much more subtly: interpretation of the whole and of the parts give the same answer. One has to decide whether to interpret directly or factor (as functions), interpret, and combine with AND/OR. It is a theoretical point about interpretation-preserving operations, that we will leave here except to say that we need not worry about deriving conflicting interpretations. 

It is natural to ask whether there is some correspondence between these subdivisions of the model and subdivisions of the data. After all, a parsimonious model should only apply differing reasoning to differing cases. This correspondence indeed exists. 

\begin{definition}\label{def:MmSpace}\MmSpace{}:  \\
Let $\MmFun{}$ be a \MmRep{} of $\net$ with $\MmFactor{}s$  $\MmPart{}_1,\ldots,\MmPart{}_k$. For each \MmFactor{}, $\MmPart{}_j$, let $\{x | \MmPart{}_j(x)=\net(x) \}$ be the corresponding \MmSpace{}.
\end{definition}

Note that these spaces partition the input: because Min (resp. Max) agrees with at least one of its inputs at every point, then so does $\MmFun{}$, which is a finite combination of the two. Therefore, each ECG falls into some \MmSpace{}, and we refer a collection (e.g. the training set) of ECGs all belonging to the same one a \MmConcept{}. Note also, that on this subset of, the \MmFactor{} and $\net$ are the same function, so interpreting the former is equivalent to interpreting the later conditional on this additional information. While, to our knowledge, this section is novel, in the next section we need to briefly dip into the background material to borrow a math technique. 






\subsection{Discussion and Approach}\label{sec:approach}
The section discusses \Mm{} representations of a neural network in theory, in the literature, and in our approach. By a neural network, we mean recursive composition of $d$ ``layers", each of which is an affine function following a ReLU function,  $\relu(x)_i=\max\{0,x_i\}$, the output of each usually being referred to as an ``activation". To build an example around $1$ layer, let us denote by $z(x)$ or simply $z$ the last activation (that is not the output), so that
\begin{align*}
    \net(x)=\B{d}+\W{d}\relu( z(x) ). 
\end{align*}
Here $\B{d}$ and $\W{d}$ are the bias and linear components affine map in the last of $1,\ldots, d$ layers. A helpful approach is to split the sign components of any vector or matrix, $M$, by using the corresponding subscript, $(M_{\pm})_{i,j}=\max\{0,\pm M_{i,j}\}$. The idea is to organize terms in the optimization so that the \textit{greedy} choice for each $\relu$ linear component agrees with the one actually realized by the network. Continuing our example we have,

\begin{align*}
    \net(\x)=\B{d}+\max_\mu \W{d}_+\mu( z(x) ) - \max_\tau \W{1}_-\tau( z(x) )
\end{align*}

Here, we are considering $\mu$ and $\tau$ to be optimized over binary diagonal matrices--simply enough, they are always driven to ``zero out" any negative components. They optimize different variables so, trivially, a difference of maxima can be written equivalently as a MaxMin or a MinMax of the difference, which in this case is a linear function of $z$. All this so far is common to both \citep{Zhang2018} and \citep{Snyder2020a}, but at this point they give qualitatively different approaches to multi-layer networks. 

For his original interest in that class of functions, \citep{Zhang2018} says \textit{any} neural network can be written as a difference of maxima using only linear functions of the input. At first this sounds good. We did not even ask for each \MmFactor{} to be interpretable, let alone linear. But, something has to give. If you design your \MmFactor{}s to be linear, then it will taken very many of them to represent $\net$. If interpretable functions are ``closed under composition", then the \MmRep{}, $\MmFun{}$, will be too complicated a to be useful. 

As an alternative, we follow the layer-wise approach taken in \citep{Snyder2020a}. Specifically, we are following Algorithm 2 in the appendix. We will give a quick summary in our notation. The idea is to simply recurse the 1 hidden layer expansion we demonstrated. 
In the first step $\MmFun{}$ has $MaxMin$ structure and arguments $\MmPart{}_1,\ldots,\MmPart{}_k$ that are $d-1$ layer neural networks. Only the first $d-2$ layers of these networks are identical to the original. 
If we treat each $d-1$ layered network individually in the same fashion as the original, then we get nested MaxMinMaxMin structure for $\MmFun{}$ which optimizes over terms that are each $d-2$ layer functions. We continue recursively. The indices and number of functions grows like the number of linear regions achieved in the terminal layers. 

However, we cannot apply this method exactly because \citep{Snyder2020a} only outline the approach for fully-connected (FC) layers, while in our setting 1D convolutional (Conv) layers and Max Pooling layers (MP) are standard. These layers \textit{can} definitely be used in a similar scheme, but we found it simpler to restrict Conv and MP layers to the initial stages, so that the factorization only ``sees" the later FC layers. Because Conv and MP layers can also be represented by FC networks, the algorithm cannot tell which has generated the \MmFactor{}s and as such still functions correctly.

\section{Methods}
\begin{figure}[h!]
  \centering 
  \includegraphics[width=5in]{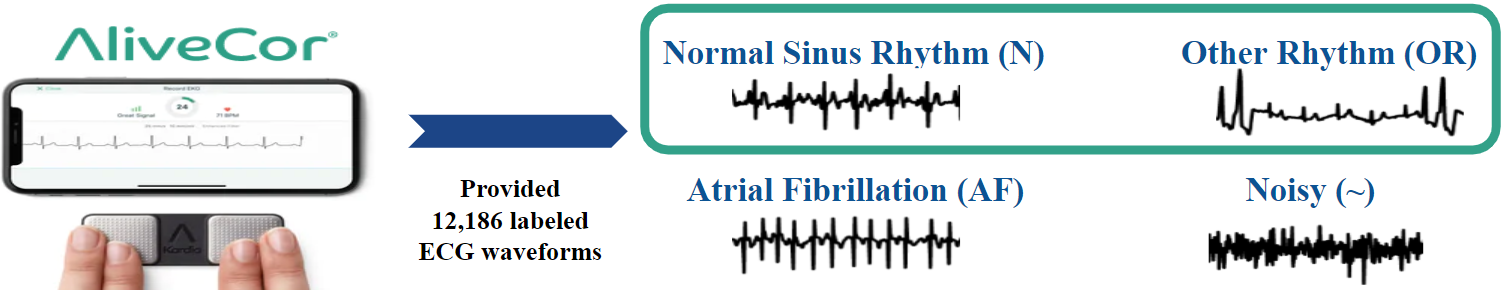} 
  \caption{The PhysioNet 2017 Dataset.}
  \label{fig:data_generation} 
\end{figure} 
This section defines an experimental design that reflects the aims, concepts, and techniques from previous sections. 
Here the largely theoretical exposition turns sharply practical, as we detail the actual physical steps and procedures to produce our experimental outputs. These include dataset creation, network design, training protocol, as well as our network-\Mm{} Conversion algorithm and supporting heuristics. 


\subsection{Dataset and Data Preprocessing}

We used ECG waveform data from the PhysioNet 2017 Computing in Cardiology Challenge
\citep{Clifford2017}, 
which was also a component of \citet{Goldberger2000}. The challenge encouraged development of algorithms that differentiate single-lead ECGs labeled as atrial fibrillation (AF), normal sinus rhythms (N), other rhythms (OR), and rhythms too noisy for classification ($\sim$). While the PhysioNet dataset is often used for bench-marking classification models, we are instead interested in demonstrating the \textit{interpretation} of a classification model. To facilitate this study, several simplifications were made to the original classification task.
\begin{figure}[h]
  \centering 
  \includegraphics[width=\linewidth]{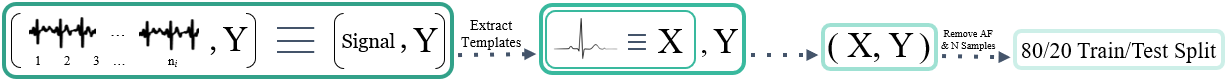} 
  \caption{Dataset Preprocessing.}
  \label{fig:preprocess} 
\end{figure}

The PhysioNet dataset was obtained and donated by AliveCor. Lead I (LA-RA) equivalent ECG recordings were generated using an AliveCor hand held device. Each recording ranges from $9$ to $61$ seconds. The complete dataset includes $12,186$ recordings that were partitioned in a $70/30$ split, resulting in a training set of $8,528$ and a test set of $3,658$.  
For our datasest, the recordings with ($\sim$) and AF labels were removed. 
As we have access to only the training set, we perform an additional $80/20$ split at random to generate our train and test data. The PhysioNet train/test split was completed along waveform lines to prevent patient data from belonging to both the training and test set. Instead, our model inputs consist of short snippets of ECGs called ``templates". For simplicity here, the patient information was discarded. The $R$ waveform of each template is aligned, and light filtering is performed. Each template ``inherits" the label pertaining to the waveform it derived from, as if each ECG complex within the waveform exhibits that labeled morphology. 

The data distribution samples uniformly a (waveform,label) pair from either the train or test set, and subsequently samples uniformly a template or ECG complex from that waveform. The DNN model is trained to minimize the negative log likelihood of the label given the template.




\subsection{Architecture Design}
\begin{figure}[htbp]
  \centering 
  \includegraphics[width=4.5in]{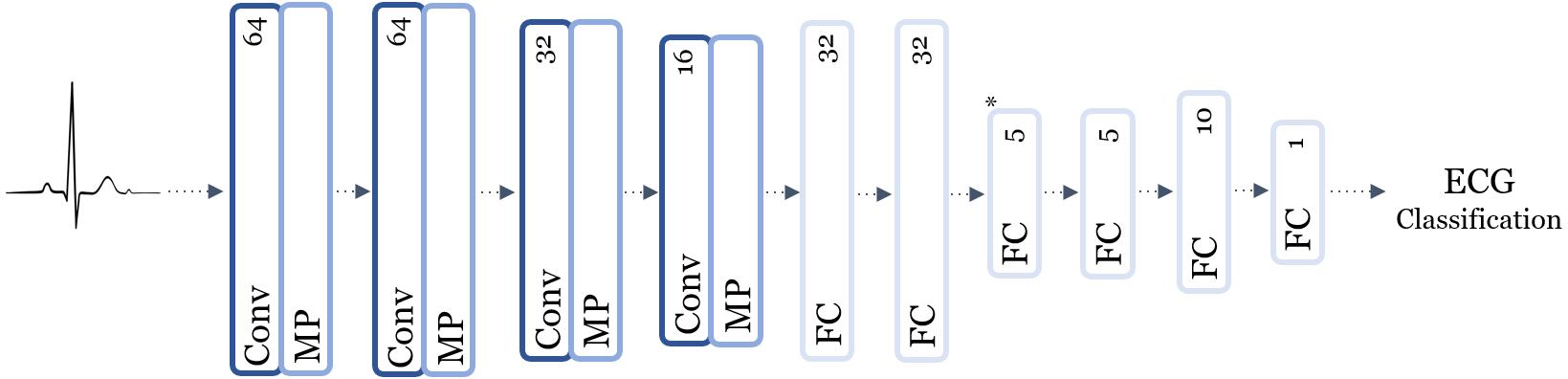}
  \caption{Architecture of the network. 
  For the convolutional layers (Conv), we use kernel sizes of $6$ and $4$ for the first and second halves, respectively, and we use strides of $2$. The max pooling (MP) layers all had pool sizes of $2$ and strides of $1$. Final layers of the neural network were fully-connected (FC).} 
  \label{fig:arch}
\end{figure}

We used a convolution layer model roughly based on the one in \citep{Goodfellow2018} but with some adaptations particular to our setup. Overall, the network consisted of several convolutional alternating convolution and max pooling layers, followed by several fully-connected layers. Layers aside from the max pooling and terminal layers were followed with a ReLU nonlinearity. An illustration is shown in Figure \ref{fig:arch}. 

We reduce number of convolutional filters and degree of pooling to reflect the change from whole ECG inputs to shorter waveform inputs. The size of fully-connected filters in the later layers was also reduced (without more than 2-3\% change in model accuracy) to reduce the number of linear pieces comprising the terminal 4 layers. 

%
%
%
%

%
\subsection{Model Training}
Neural network training was done with the Tensorflow library. None of the values were tuned, and most were simply inherited from previous reused code. We used the Adam to optimize a sigmoid cross-entropy loss with $1e^{-5}$ learning rate and batch size of $64$. We trained for $80$ training epochs for the models in this paper, but we have no evidence that this long length of time is necessary or important. 


\subsection{Calculating \MmRep{}, \MmConcept{} Partitions}

This section covers unique implementation details. Definitions and algorithm for calculating \MmRep{} are given in Section \ref{sec:approach} and \citet{Snyder2020a}, Algorithm 2. By \MmConcept{}s, we refer to the \MmSpace{}s, defined at the end of Section \ref{sec:motivation} and restrict them to training samples.

We apply these algorithms proposing our ``input" is actually the embedding output from the first $5$ neuron layer, indicated by the asterisk (Fig. \ref{fig:arch}). 
The complexity of this approach as implemented grows roughly with the number of linear regions, which is kept reasonably small ($10-100$) by the smaller width. 
Like \citet{Snyder2020a}, we identify these regions defining \MmRep{} and \MmFactor{}s using a grid search. Ours are unbounded potentially, so we use $99^{\text{th}}$ percentiles.  

Min and Max, being differences of maxima, commute and thus provide a choice whether Min or Max should lead each layer representation. We lead with Min. The motivation is that, since we classify Normal (positive) vs Other (negative) rhythms, we want to allow for AND to be the highest level interpretation. The goal is to reach an interpretation like, ``$x$ is Normal"  iff ``$x$ not diagnosis 1", AND ``$x$ not diagnosis 2", etc. 

The \MmConcept{}s can be conveniently calculated alongside the recursion building the \MmRep{}. At each step, simply divide the ECGs associated to the current component to the ArgMax/Min of the substituting representation. An important note is that there were some \MmFactor{}s with empty \MmConcept{}. This is partially because the grid search may explore regions that inputs do not, but also because the \MmRep{} is only guaranteed to be correct, not minimal. We drop these from the visualization explained next section. 

The most obvious way to interpret each \MmFactor{} turns out not to work for ECGs. If one views each \MmFactor{} as a function on the entire space, as was done in \citet{Snyder2020a} with MNIST digits, the corresponding interpretation will be correct but perhaps not the simplest correct one. Each \MmFactor{} becomes easier to interpret in the context of the others by deriving additional descriptive input classes: Small changes to each \MmFactor{} only ``cause" 
the model to change outputs on a subset of inputs we call a \MmConcept{}. 

\subsection{Interpretation through Visualization}\label{sec:method vis}

\begin{figure}
    \centering
    \includegraphics[width=4.5in]{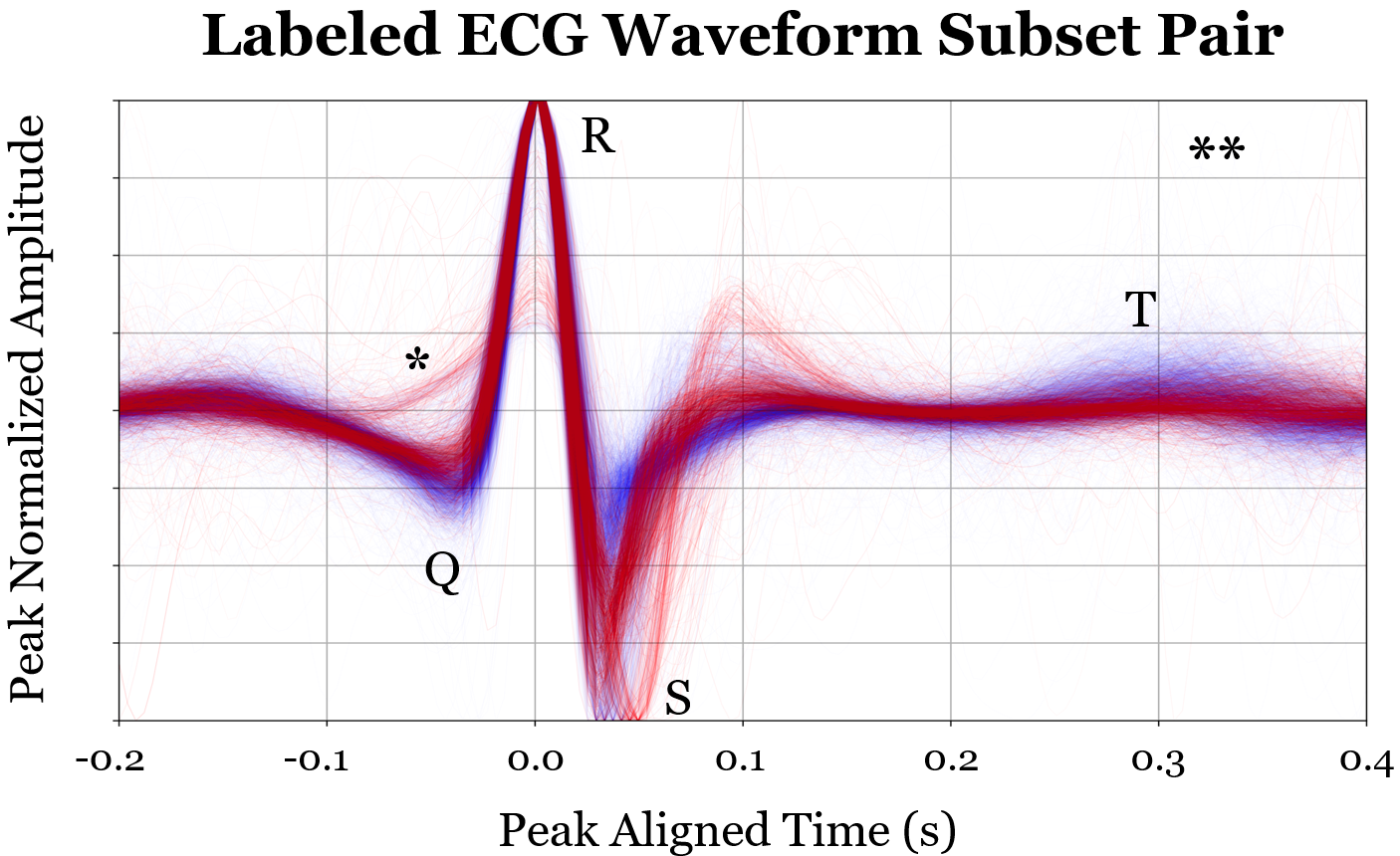}
    \caption{ECG plot with peak normalized amplitudes. The figure, a composite visual of individual ECG recordings, quickly conveys variety, distribution, and clustering of trajectories to an observer. Abnormal (negative) classes have a combination of higher \textbf{Q} waves, more depressed and extended \textbf{S} waves and absent \textbf{U} waves. These observations reveal fundamental vibrations of ECG potential that represent class characteristics. To obtain this figure, we align each waveform at $0.0$ seconds in order to closely compare them. Several thousand individual ECG recordings are then drawn with replacement from the abnormal (negative) and normal (positive) classes. The line transparency is adjusted and the ECGs are directly overlaid.} 
    \label{fig:labeledECGexamples} 
\end{figure}

While DNN functions can be difficult to visualize directly, we can characterize them through the data partitions associated with their component parts. Ultimately, we want to understand how the classifications boundaries split these characteristic sets. A ``tutorial" example of one such \MmConcept{} visualization is explained below and given in Figure \ref{fig:labeledECGexamples}.

For a waveform of a single class label, we first R-wave peak normalize and align the waveforms. Then we use alpha blending to overlay the waveforms with red and blue corresponding to label, which creates darker areas of the graph where many ECGs have the same normalized potential at the same time point. When plotting a sample of $4000$ waveforms with a small alpha value ($<0.01$), anomalies plotted by a few waveforms are hardly noticeable. Alpha and other parameters such as line thickness were chosen by visual tests to ensure the graph was not over saturated with lines. 


\section{Results} 

\begin{figure}
  \centering 
  \includegraphics[width=\linewidth]{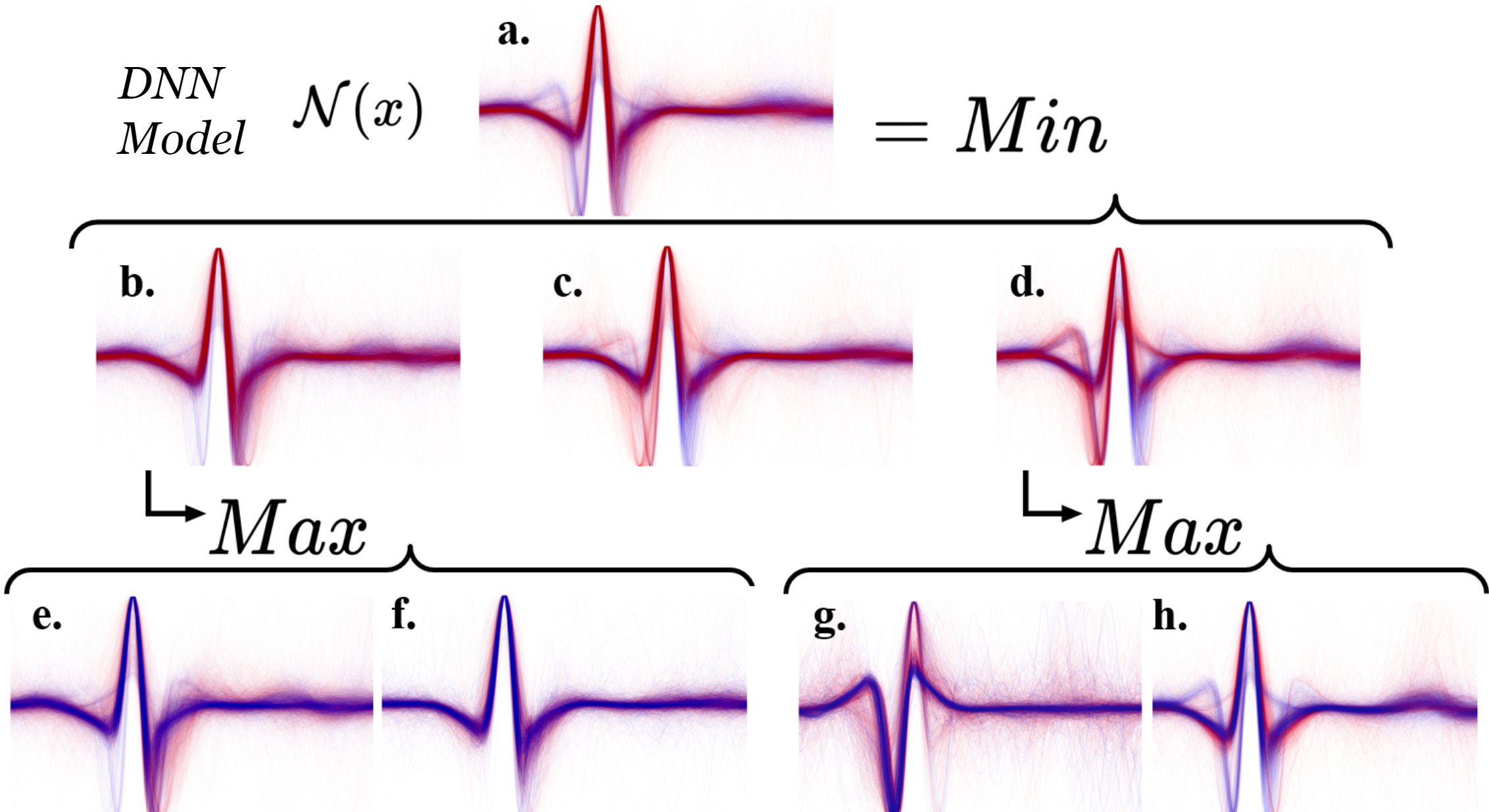}  
  \caption{A Visual Explanation of a DNN ECG model as emergent from Min,Max equations governing 
  interpretable component parts. The figure is also an equation for the neural network, parameterized by \MmFactor{}s represented by the corresponding \MmConcept{}s. Each waveform sample is drawn in \textbf{a.} and once in the ArgMax or ArgMin component visualization following each bracket. Details and analysis in text. But, many interesting features left for the reader to explore. The \MmFactor{} in \textbf{c.} has no more subdivisions at this depth, so it equally belongs among the third row.
  }
  \label{fig:the main figure} 
\end{figure}

Our model achieves $74\%$ accuracy. Other challenge models at the time achieved accuracy in the low $80^\text{th}$ percentile. Of course, diagnoses defined in terms of the the R-R interval will be harder without access to whole ECG recordings. Also, the restrictions we placed on the size of the terminal layers may have made it more difficult to classify certain patterns. But, this quite reasonable accuracy indicates our simplified model is 
\textit{qualitatively representative} of an out-of-the-box ECG model in practice. By extension, we argue our interpretation achieves this level of accuracy as well. 
When we interpret this model by deriving a component representation of the last $2$ hidden layers, we get a very rich and informative story. Correspondingly, its representation is also very information dense and needs to be digested slowly, at multiple zoom-scales, and with color. By visualizing and arranging the aggregate waveform of each \MmConcept{} using the technique discussed in Section \ref{sec:method vis}, we obtain joint interpretation of each component as it relates to the overall model. Refer to Figure \ref{fig:the main figure} throughout.

The top row is easiest to understand, and can be viewed independently of the rest. The image \ref{fig:the main figure}\textbf{a.} is a composite of every training sample as labeled by the final trained model. Equally valid would be a representation using test samples; they simply answer different questions. It is useful to compare both but beyond our scope. Instead, we want to follow a relatively simple thread. 

The reader may have noticed some of the waveforms plotted are upside-down, having their polarization inverted. The two downward extensions of the Q and S waves (we'll call them \textit{legs}) are present in most images, except some in the last row. What happened was an extremely fortuitous, informative accident. In our attempt to reproduce the code from \citet{Goodfellow2018}, we missed the portion that corrects the polarization. Depending on how peak alignment was done, the R wave was sent to either the Q or S leg. The effect of this is to artificially create additional waveform morphologies and phenotypes. This is suboptimal from the point of view of performance maximization. But in fact, it is a wonderful wrinkle--one representative of the realities of clinical modeling--that we can use to demonstrate the potential for our interpretation method. 

Surely, in practice similar mistakes occur. One usually cannot easily verify if errors exist in some clinical data samples. Notably, the model behavior does not distinguish between clinical and artificial data structure. So it is extremely important to understand how such mistakes and structures in general become represented in our models. Does the model even identify polarization inverted waveforms as a distinct \MmConcept{}s? If so, then perhaps further analysis will show it also discovers clinical diagnoses based on morphology. As it turns, the neural network model has three fundamental modes that differ with respect to how they treat inverted Q and S leg waveforms.

In the second row, Figures \ref{fig:the main figure}\textbf{b.},\textbf{c.},\textbf{d.}, we can begin to understand these modes or \MmFactor{}s. The combination of the first and second row is also an equation: \textbf{a.}$=$Min(\textbf{b.},\textbf{c.},\textbf{d.}) or with \MmFactor{}s $\MmPart{}_b,\MmPart{}_c,\MmPart{}_d$ it says $\net(x)=\text{Min(}\MmPart{}_b,\MmPart{}_c,\MmPart{}_d$). Each sample waveform is represented visually on both sides of this equation (in fact once per row). Because each column is a representation of a \MmConcept{}, each sample is only drawn in the visualization of the \MmFactor{} that achieves the minimum of $\MmPart{}_b,\MmPart{}_c,\MmPart{}_d$, determines the output class label independently of the other two. Therefore, we use the relationship between the two rows to understand the label given to a single sample: A waveform is considered normal iff it is drawn in blue in the top row iff it is drawn in blue in one of the three second row figures. These characterizations also hold between the second row and bracketed third row \MmFactor{}s. 

\def\bfa{\textbf{a.}}
\def\bfb{\textbf{b.}}
\def\bfc{\textbf{c.}}
\def\bfd{\textbf{d.}}
\def\bfe{\textbf{e.}}
\def\bff{\textbf{f.}}
\def\bfg{\textbf{g.}}
\def\bfh{\textbf{h.}}
Having defined what a row is, we can see emergent structures in the Figures \ref{fig:the main figure}\textbf{b.},\textbf{c.},\textbf{d.}. We see by the color of each Q and S leg in the second row that alignment direction of the inverted waveform, becoming either the Q or S leg, is a central organizational theme for the neural network. We see \ref{fig:the main figure}\bfa{} accounting for about half of the Q leg positive/Normal samples and all of the S leg negatives/Other samples, \ref{fig:the main figure}\bfc{} sharing about half of each of the negative Q leg and positive S leg samples.  The story with \ref{fig:the main figure}\bfd{} is similar to \ref{fig:the main figure}\bfc{}, but with mixed Q leg contributions, and complex dynamics overall. Amazingly, this complex structure in \ref{fig:the main figure}\bfd{} gets \emph{easier} to interpret the farther we carry the interpretation.

In Figures \ref{fig:the main figure}\bfg{}and \bfh{}, we observe interpretable component representation of the the \MmFactor{} in \ref{fig:the main figure}\bfd{} We discover that the model \textit{does} treat polarization inverted waveforms as a fundamentally distinct class. 
The \MmFactor{} $\MmPart{}_g$ generates the decision boundary for the inverted waveforms where $\MmPart{}_d$ is optimal among the first row. 
The \MmFactor{} in \ref{fig:the main figure}\bfh{} handles the complement of upright waveforms (along with \textit{some} inverted ones). 
Observable in both \ref{fig:the main figure}\bfd{} and \bfh{} (perhaps best in \ref{fig:the main figure}\bfh{}) are these contrasting red-blue bands contouring the QRS complex. These allow us to interpret how decisions are made among upright ECGs. The red outer R wave detailing in \ref{fig:the main figure}\bfh{} suggests a component that labels as negative those samples with R waves that rise too slowly. Likewise, we see abnormal diagnoses associated with P waves that are too deep, and Q waves rising too slowly. 

An important point to remember is that these interpretable structures are in no way obligated to manifest. Each sample must be present somewhere in each row, but we they do not also need to be organized and sorted in a way that seems reasonable and interpretable to us. As the complexity of the functions, for example, $\MmPart{}_b,\MmPart{}_c,\MmPart{}_d$, is not controlled, these samples could take any arrangement with sufficiently expressive lower layers. A second point: even if they stay interpretable, we don't know of any theoretical maxim that says these interpretations should so quickly become simpler.

\section{Discussion}
We all have a mental image of what clinical relevance ``looks like". Perhaps, one recalls previous papers that tried to solve similar problems. Why does this one introduce so much unorthodoxy and detail so as to obscure that clinical connection? Let us try to motivate why something in the this style of approach is really prerequisite for continuing to make intentional forward progress with Deep Learning, in particular in medicine. 

Consider the ``stadium wave", in which successive, adjacent groups of seated spectators of sport stand and raise their arms upwards. What neural networks do really well is generate high level concepts by mapping low level inputs, such as pixels, sound amplitudes, heart rates. Interpreting these end-to-end is impossible. Doing so would be analogous to trying to interpret a neural network that predicts the frequency of ``stadium wave" behavior from the mere DNA sequences of the sport spectators, without pausing to understand how the model represents ``humans" as a concept. It becomes much easier to understand, control, debug, iterate, interpret, and learn from the model if you can consider the wave behavior model from the perspective of changes to stadium members' behavior, rather than individual nucleotide base-pairs. Component interpretations are important for studying modeling with Deep Learning, just as cells are important for studying medicine with humans. 

This is especially relevant for medicine where we anticipate these model components can have interpretations that circle back and, in turn, teach us about medicine. We have demonstrated this is exactly what occurs with our approach in Figure \ref{fig:the main figure}, where phenotype corresponds to morphology. If there are natural clusters that models routinely find useful for modeling vast quantities of data that humans simply do not have the lifespan to access, then these are useful targets for follow up studies to try to find a common physiologic mechanism.

We would assess this method as not ready for direct use by clinicians but in need of interest cultivation and  improvement to supporting algorithms. It works, but it's fragile. One has to properly contextualize: when deep learning paper publishes a model, that work is subsided by \textit{decades} of experience, supporting algorithm development, and standardized libraries. Because our approach is genuinely new, we lack all of that again
\footnote{Even the visualization implementation deserves its own further study. To say nothing of the complex subjective human perceptions, we need a custom rendering to blend these better, because existing software will only blend one plot at a time, give a ``painted over" feel.}. 
But we also have opportunities to improve our results by at really every step. 

\paragraph{Removable Limitations}
These are conditions we required experimentally that we believe strongly could be removed with additional theory. 
For example, we only expanded the fully-connected layers, treating the convolutional ones as an embedding. This is convenient, but unnecessary since they can always be viewed as special cases of fully-connected ones. 
Generally, one should expect the \textit{theory} to be adaptable to any piece-wise linear operation, including max-pool layers for example. 
Though, the experimental behavior properties may differ, in part because the parameterization determines the training dynamics and thereby affect the final structure.
For now, the trickiest part is keeping the \Mm{} expansion small enough when the number of neurons in the layers is very large. We accomplished this by having very small width in the later layers. 
This keeps the expansion small because there are fewer neuron state(on/off) combinations. 
We suspect that in these cases some small subset is usually sufficient to agree with model behavior with high probability. 
But, in general when this is possible is determined by the experimental data. 
We don't expect most data in high dimensions to have a circular decision boundary with small margin, but a fine approximation to a circle with many pieces would break this part. 
\paragraph{Intrinsic Limitations}
For any of this to work, interpretable component representations (1) have to exist and and (2) have to be representable to humans in some intelligible way. Unfortunately, we don't know how to substantiate either of these with theory. The former seems to happen whenever we can contrive the latter. But, it's just not clear how much we're \textit{really} asking for with that first word ``interpretable" components.

\section{Acknowledgements}
This work was supported by NSF under grants 1731754 and 1559997.

\clearpage

\bibliography{ECG-Proj}



\end{document}